\pdfoutput=1
\documentclass[11pt]{article}

\usepackage{emnlp2021}
\usepackage{preamble}

\title{Disentangling Generative Factors in Natural Language with Discrete Variational Autoencoders}

\author{Giangiacomo Mercatali
\thanks{\ \ name.surname@postgrad.manchester.ac.uk} \\
  University of Manchester  \\
  \\\And
  Andr\'e Freitas
  \thanks{\ \ name.surname@manchester.ac.uk} \\
  Idiap Research Institute \\
  University of Manchester \\
}

\begin{document}
\maketitle
\begin{abstract}
The ability of learning disentangled representations represents a major step for interpretable NLP systems as it allows latent linguistic features to be controlled. Most approaches to disentanglement rely on continuous variables, both for images and text. We argue that despite being suitable for image datasets, continuous variables may not be ideal to model features of textual data, due to the fact that most generative factors in text are discrete. We propose a Variational Autoencoder based method which models language features as discrete variables and encourages independence between variables for learning disentangled representations. The proposed model outperforms continuous and discrete baselines on several qualitative and quantitative benchmarks for disentanglement as well as on a text style transfer downstream application. 

\end{abstract}

\section{Introduction}

%% why disentanglement is relevant
A fundamental challenge in Natural Language Processing (NLP) is being able to control generative factors for text, such as tense, gender, negation, which are characterised by an entangled representation in traditional neural networks, making it difficult to control them. Disentangled representation learning aims to provide an interpretable representation of latent features, and a framework for controlling the change of specific features, by separating distinct generative factors in the data~\citep{bengio2013representation}.

%% approaches to disentanglement
Various disentanglement approaches for neural networks have been proposed. \citet{chen2016infogan} achieve disentanglement with a Generative Adversarial Network (GAN), by maximizing the mutual information between latent variables and generated samples, while~\citep{higgins2016beta,burgess2018understanding,kim2018disentangling} fine-tune the parameter which controls the KL divergence in a Variational AutoEncoder (VAE)~\citep{kingma2014auto}. In the NLP domain, \citet{john2019disentangled} use adversarial losses to separate the style and content embeddings, while \citet{cheng2020improving} disentangle style and content embeddings by minimizing their mutual information.

\begin{figure}[t]
 \centering \includegraphics[scale=1]{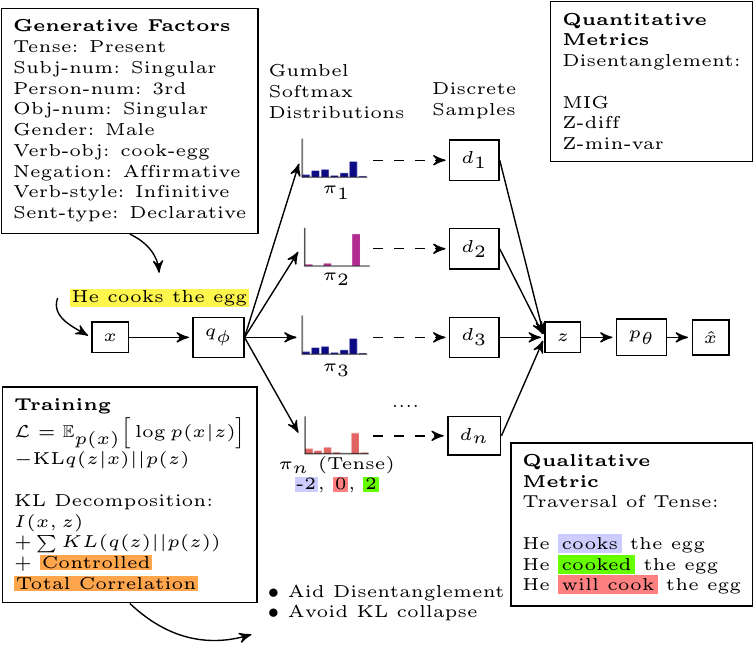} \caption{Overview of the proposed Discrete Controlled Total Correlation (DCTC) model: The KL decomposition encourages independence between variables, which are encoded as discrete latents, to capture the dimensions of linguistic features. The representation is probed with latent traversals and quantitative metrics.} \label{fig:model}
\end{figure}

% issues in NLP disentanglement: 1) countinuous encoding  2) lack of evaluation tools
After a thorough review of the literature, we find that there are currently two main issues in the area of disentanglement in NLP. First, the mentioned approaches operate using Gaussian distributions in a continuous space. Although a continuous representation may be suitable to encode images, for text data one should rather specifically consider discrete distributions of the feature set. In fact, generative features of a sentence mostly belong to a discrete domain, for example, one would encode the gender feature either as male or female, while the tense can be modelled as having three values, such as present, past, future. Secondly, while in the realm of Representation Learning, numerous quantitative and qualitative evaluation methods for disentanglement have been designed, there is still a gap in the adaptation and adoption of these methods into the language domain. As a result, despite being a critical step for enabling the interpretability of NLP models, text disentanglement is still not fully explored.

% discrete nature of language features
\noindent We address the highlighted problems as follows:

% model
\noindent 1) We design the first VAE-based architecture (shown in Figure~\ref{fig:model}), where linguistic features are encoded as discrete latent variables via the Gumbel-Softmax trick~\citep{jang2016categorical} and disentanglement is jointly enforced in the objective function. We derive a decomposition of the VAE evidence lower bound (ELBO), where independence between variables is encouraged by fine-tuning the Total Correlation term, with the goal of achieving disentanglement. At the same time, an information channel mechanism is introduced, and increased incrementally during training, to avoid the issue of the posterior collapse~\citep{bowman2016generating}.

% evaluation
\noindent 2) We provide, to our knowledge, the first extensive evaluation of the text representation, under the lenses of contemporary disentanglement methods. We propose to probe the quality of the representation by traversing and decoding each latent, expecting a disentangled representation to show only a single dimensional change (for example in tense, as in Figure~\ref{fig:model}). In addition, we show that the proposed model outperforms numerous baselines on three quantitative disentanglement metrics from the Representation Learning literature. Finally, we show the beneficial effect of the proposed representations in the task of text style transfer.

\section{Background}

\paragraph{Language VAEs.}
Language VAEs~\citep{bowman2016generating} are generative models widely used in NLP tasks such as text style transfer~\citep{john2019disentangled} and conditional text generation~\citep{cheng2020improving}. The VAE architecture consists of a decoder network $p_\theta$ and encoder network $q_\phi$, acting over a sequence of sentences $x_1, \dots, x_n$.

The VAE considers a multivariate Gaussian prior distribution $p(z)$ and generate a sentence $x$ with the decoder $p_{\theta} (x|z)$. The joint distribution for the decoder is defined as $p(z) p_\theta (x|z)$, which, for a sequence of tokens $x$ of length $T$ result as $p_\theta (x|z) = \prod_{i=1}^T p_\theta (x_i | x_{<i}, z)$.

The framework's objective is to maximize the expectation over the full dataset of the log-likelihood, in other words, $\mathbb{E}_{p(x)} \log p_\theta(x)$. However, this term is intractable, and thus the variational distribution $q_\theta$ is introduced to approximate $p_\theta (z|x)$. As a result, an evidence lower bound $\mathcal{L}_\text{VAE}$ (ELBO) where $\mathbb{E}_{p(x)} [\log p_\theta(x)] \geq \mathcal{L}_\text{VAE}$, is derived as follows:

\begin{align} \label{eq:vae_loss}
\mathcal{L}_\text{VAE} = &
\underbrace{\mathbb{E}_{q_\phi(z|x)} \Big[ \log p_{\theta} ( x | z ) \Big] }_{\numcircledmod{i}} \\ \nonumber
& \underbrace{ - \text{KL} q_\phi(z|x) || p(z) }_{\numcircledmod{ii} }
\end{align}

In Eq.~\eqref{eq:vae_loss}, \numcircledmod{i} represents the reconstruction error, which encourages the model to encode the data $x$ into the latent variables $z$, while \numcircledmod{ii} represents the KL divergence regularization term, which forces the variational distribution $q_\phi(z|x)$ to be similar to the prior $p(z)$.

\paragraph{Disentanglement in VAEs.}
The key intuition of a \textit{disentangled} representation is that it should separate the independent factors of variation in the data. As a result, a change in a single underlying factor is reflected into a change in a single factor of the learned representation, while being invariant to changes in other factors~\citep{bengio2013representation}.

Early approaches to disentanglement are GAN-based~\citep{chen2016infogan}, however, most contemporary methods focus on the VAE framework, which improves the training stability. The $\beta$VAE framework~\citep{higgins2016beta} introduces a hyperparameter $\beta$ for tuning the KL term \numcircledmod{ii} in the ELBO~\eqref{eq:vae_loss} (i.e. $\mathcal{L}_{\beta\text{VAE}} = \numcircledmod{i} - \beta \numcircledmod{ii}$) and demonstrates that on one hand, a $\beta>1$ leads the model to learn disentangled representations, however, the disentanglement comes at the cost of a low reconstruction fidelity.

Recent approaches~\citep{kim2018disentangling,chen2018isolating} demonstrate that disentanglement can be achieved without sacrificing the reconstruction fidelity, by decomposing the KL term in the ELBO, and tuning only the terms that encourages independence between latents, while not penalizing the mutual information between latents and data.

%The tradeoff of $\beta$VAEs is further explained from~\citet{kim2018disentangling}, that decompose the KL term \numcircledmod{ii} in Eq.~\eqref{eq:vae_loss} into the mutual information between the latents and the data $I(x,z)$, and show that penalizing $I(x,z)$ reduces the amount of information about $x$ stored in $z$, leading to poor reconstruction.

% Discrete TC VAE
\section{Proposed Approach}

\paragraph{Model Overview.}
Differently from previous approaches to disentanglement~\citep{higgins2016beta,kim2018disentangling,chen2018isolating}, we focus our efforts into leveraging the discrete generative factors present in natural language, and design a framework, which we name Discrete Controlled Total Correlation (DCTC), where language factors are encoded as discrete latent variables, while the representation is enforced to be disentangled. We first reformulate the continuous VAE for discrete variables by providing a suitable discrete reparameterization process. We then decompose the ELBO, and provide a modification of the Total Correlation term for controlling disentanglement.

\subsection{Linguistic Features as Discrete Variables}
\paragraph{Discrete Variables.}
Discrete representations present strategic advantages for modeling natural language. Firstly, a discrete encoding can capture the dimensions of a linguistic feature (e.g.\ tense is a three values variable). In addition, since the dimensions of language features are known, it is possible to induce semantic factor biases from them, and thus enhance disentanglement~\citep{locatello2019challenging}.

In order to model the discrete generative factors of a sentence or passage, we first define a set of discrete latent variables $d$, that are sampled from a Gumbel-Softmax distribution~\citep{jang2016categorical} and a posterior $q_\phi(d | x)$. This is the first step to encode the discrete factors of variation in natural language, such as tense and gender. The continuous VAE in~\eqref{eq:vae_loss} becomes:
\begin{align}
    \label{eq:discrete_vae}
    \mathcal{L} = & \mathbb{E}_{q_\phi(d|x)} \Big[ \log p_{\theta} ( x | d ) \Big] \\ \nonumber
    & - \text{KL} \infdiv{q_\phi(d | x )}{p(d)}
\end{align}

\paragraph{Discrete Reparameterization.}
In order to back-propagate with discrete variables, we need to extend the reparameterization-trick from~\citep{kingma2014auto}. In fact, the vanilla VAE only considers a continuous variable $c$, and $q_\phi(c|x)$ is parameterized by $q_\phi(c|x) = \prod_i q_\phi(c_i|x) $ where each distribution is Gaussian. Since in our loss~\eqref{eq:discrete_vae} we consider $d$, that is a discrete variable, $q_\phi(d|x)$ is non-differentiable, thus we resort to the Gumbel-Softmax trick~\citep{jang2016categorical}, which provides a tool for sampling from a continuous approximation of a discrete distribution. The Gumbel-Softmax trick considers a discrete variable with class probabilities $\pi_1, \dots, \pi_k$, and draws samples $g_1, \dots, g_k$ from a Gumbel distribution, as follows:
\begin{align}
 y_i = \frac { \exp( (\log(\pi_k) + g_k) / \tau)} {\sum_{j=1}^k \exp((\log_{\pi_j} + g_j)/\tau)}
 \label {eq:gumbel_soft}
\end{align}

By plugging in the samples $g_i$ and the class probabilities $\pi$ in Eq.~\eqref{eq:gumbel_soft}, we generate a $k$-dimensional vector $y$, that is the continuous approximation of the one-hot-encoded representation of the discrete variable $d$. In fact, as $\tau$ approaches 0, samples from the Gumbel-Softmax distribution become one-hot, making it discrete.

With this approximation mechanism in place, we can define the prior $p(d)$ in Eq.~\eqref{eq:discrete_vae} as a product of Gumbel-Softmax distributions, which makes the decoder $q_\phi(d|x)$ differentiable and enables us to train the discrete VAE. For our purpose, each distribution represents a linguistic feature, and can be set to a discrete dimension.

\paragraph{Model Architecture.}
The model architecture is depicted in Figure~\ref{fig:model}. We consider a training sample such as a sentence $x$ of length $T$ composed by $x_1, \dots, x_T$. Our model is built using a LSTM encoder which receives a sentence $x$, and samples $n$ discrete latent variables; and a LSTM decoder which receives $n$ discrete variables and merges them in a sentence. Each discrete latent variable $\pi_{i}$ is sampled from the Gumbel-Softmax distribution, then the Gumbel-Softmax trick~\citep{jang2016categorical} is used for drawing a discrete sample $d_i$. Finally, the samples $d_i,\dots d_n$ are fed to the decoder after being concatenated. The proposed model also aims to learn disentangled representations of sentences. This is achieved with a modification of the ELBO, where we encourage independence between latent variables, as explained in the following section.

\begin{figure*}
$
\mathbb{E}_{p(x)} \Big[ \text{KL}  q_\phi(d |x) || p(d) \Big] =
\underbrace{\text{KL}  q(d ,x) || q(d)p(x) }_{\numcircledmod{1} }
+ \underbrace{\textstyle{\sum_j} \text{KL} q(d_j)||p(d_j )}_{\numcircledmod{2} }
+ \underbrace{\gamma \Big| \text{KL} q(d) || \textstyle{\prod_j} q(d_j) - C_{d} \Big| }_{\numcircledmod{3} }
$
\caption{Discrete Controlled Total Correlation decomposition. The term \numcircledmod{3} encourages disentanglement.}
\label{eq:elbo_dec}
\end{figure*}

\subsection{Controlled Total Correlation}

\paragraph{ELBO Decomposition Design.}
For creating our objective function, we consider the KL term in Eq.~\eqref{eq:discrete_vae} in expectation over the data, and decompose it, guided by the following considerations. On one hand, we aim to include a Total Correlation (TC) term in our decomposition, because TC is a measure of dependency between variables, and thus, a penalty on it may force the model to find independent factors in the data and strongly aid disentanglement~\citep{kim2018disentangling,chen2018isolating}. On the other hand, a penalty on a TC term may lead to a KL vanishing issue~\citep{bowman2016generating}, because it may cause the decoder to ignore the information stored in the latent encoding, leading to poor reconstruction fidelity. Thus, a mechanism to avoid this issue should be integrated in the decomposition.

As a result, we decompose the KL to obtain a TC term, for aiding independence, and enhance it by introducing two parameters. First, we add a discrete information capacity $C_{d}$, which controls the amount of information that is passed into the TC term. The idea is that by increasing the information channel gradually, starting from zero, we tackle both the KL collapse issue and the high reconstruction loss of the $\beta$VAE. Furthermore, we introduce a variable $\gamma$, to enforce the TC term to match the discrete information capacity $C_{d}$. The final decomposition is shown in Fig.~\ref{eq:elbo_dec}, where $d_j$ is the j-th dimension of the latent variable $d$.

\paragraph{Analysis of Components.}
The first term \numcircledmod{1} in Fig.~\ref{eq:elbo_dec} is known as the index-code Mutual Information~\citep{hoffman2016elbo}, and it represents the mutual information between the data and the latent variable. This term has been studied in various disentanglement models, for example~\citep{chen2016infogan} claim that high mutual information is beneficial for disentanglement, and similarly, \citet{zhao2019infovae} propose the dropping of the penalty on this term to aid disentanglement. However, this notion is not universally accepted, as~\citep{burgess2018understanding} showed that a penalty on \numcircledmod{1} can also encourage disentanglement.

The second term \numcircledmod{2} prevents latent dimensions from deviating from their priors. However, this term does not have theoretical properties that suggest its utility in the enhancement of disentanglement. For the above reasons we don't fine-tune the first two terms of the decomposition and focus on the third, which we find to be the most influential for enforcing disentanglement in terms of information theoretic properties.

We name the third term \numcircledmod{3} the Controlled Total Correlation (CTC). In its original form, the Total Correlation~\citep{watanabe1960information}, is given as $\text{KL} \infdiv{q(z)}{\overline{q}(z)}$, where $\overline{q}(z)=\prod_j q(z_j)$, and it represents a generalization of the mutual information, which measures the dependence between variables. Our variation of the TC allows us to control the amount of information encoded in the discrete channel, thus avoiding the collapse of the term.

\paragraph{Training Procedure.}
To obtain the final loss function of the proposed Discrete Controlled Total Correlation model (DCTC), we consider the KL decomposition described in Fig.~\ref{eq:elbo_dec} and replace it in the original KL term in the discrete VAE ELBO in Eq.~\eqref{eq:discrete_vae}. The final loss function for our DCTC model is reported in Eq.~\eqref{eq:dtc_vae}:
\begin{align}
 \label{eq:dtc_vae}
 \mathcal{L} =
 \mathbb{E}_{q_\phi (d|x)} \Big[ \log \ p_\theta(x|d) \Big]
  - \numcircledmod{1} - \numcircledmod{2} - \numcircledmod{3}
\end{align}

During training, we increase gradually the discrete information capacity $C_{d}$ while keeping $\gamma$ fixed. On one side, this linear increase tackles the known issue of KL vanishing for text VAEs~\citep{bowman2016generating}, on the other side, it let the model maximise disentanglement. More specifically, for low values of $C_{d}$ the TC term is collapsed to zero, and the reconstructed sentences are not faithful. However, as $C_{d}$ is increased, the TC terms start to become greater than zero, and the reconstruction becomes more accurate. This improvement continues until the TC factors for each discrete latent are non-zero, and the reconstruction is identical.

\paragraph{Parameter choice.}
The choice of the parameters $\gamma$ and $C_{d}$ is derived from experimental results and it is guided by some necessary constraints. First, there should not be a tradeoff between a smaller reconstruction error and the information capacity constraint, thus the value of $\gamma$ needs to be large enough to maintain the capacity at the desired value. Second, after constraining the capacity of the discrete information channel, $C_{d}$ can be chosen to maximise the capacity of the channel, as a result, the model is prompted to use all the latent variables of the discrete distribution.

\section{Related Work}
%Our work relates to disentanglement approaches in NLP and to approaches modeling discrete latent variables, and brings them together in a novel contribution.

\paragraph{Disentanglement in NLP.}
We identify two types of approaches. 1) Multiple-losses: \citet{hu2017toward,john2019disentangled} encourage disentanglement with adversarial losses for style transfer, while \citet{sha2021multi} propose to improve the training stability, using multiple non-adversarial losses. 2) Information-theoretic: \citet{cheng2020improving} propose to disentangle style and content by minimizing the mutual information between the latent and the observed variable, while \citet{colombo2021novel} introduce an upper bound of mutual information, showing its benefits in fair classification. Differently from all these approaches, we model linguistic features as discrete variables, which allows us to enforce control on the encodings' dimensions.

\paragraph{Discrete Latent Variables.}
Various approaches present discrete encodings for language generation. \citet{shu2020controllable} enhance control and diversity in generation with latent spaces, while \citet{guo2020evidence} leverage textual evidence to guide the generation. Both methods are based on the VQVAE~\citep{van2017neural}, which we consider in our experiments. \citet{bao2020plato} encodes discrete latent variables into Transformer blocks for dialogue generation. Differently from these methods, we leverage discrete variables to optimize disentanglement.

\section{Experiments}
In this section, we evaluate the disentanglement of the proposed model with qualitative and quantitative methods, against several baselines. Furthermore, the benefits of our model's encodings are demonstrated in the downstream task of text style transfer.

\subsection{Qualitative Evaluation}

\paragraph{Latent Traversals.}
After training our model, we can evaluate the disentanglement quality of the representations by analysing the traversals of the latent space. Traversal evaluation is a standard procedure with image disentanglement~\citep{higgins2016beta,kim2018disentangling}, but represents a novelty for text datasets.

\begin{figure}[tp!]
\subfloat{ \includegraphics[clip,width=\columnwidth]{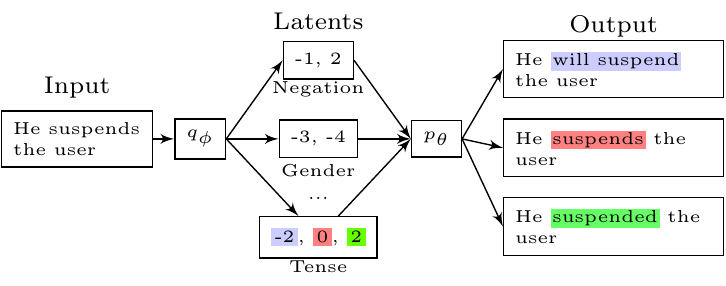} } \\
\subfloat{ \includegraphics[clip,width=\columnwidth]{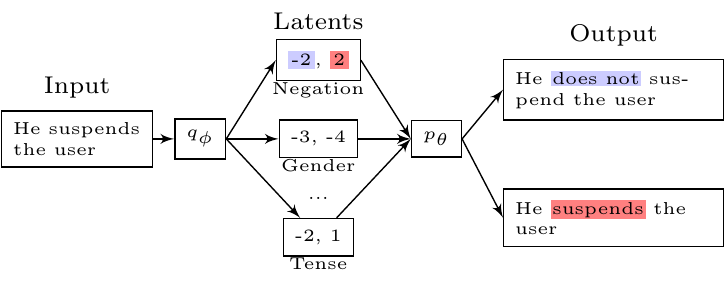} }
\caption{Traversals of generative factors. Top: tense. Bottom: negation.}
\label{fig:traversal}
\end{figure}

A visual explanation of how a traversal for textual data works is provided in Fig.~\ref{fig:traversal}. The traversal of a latent factor is given by decoding the vectors corresponding to the latent variables, where the evaluated factor is changed within a fixed interval (e.g.\ [-2, 2]), while all others are kept fixed. If the representation is disentangled, when a latent factor is traversed, the decoded sentences should only change with respect to that factor. This means that after training the model we are able to probe the representation for each latent variable.

\paragraph{Experimental Setup.}
We evaluate the latent traversals on the dSentences dataset~\citep{m2018deep}, which is composed by 37,000 sentences, and provides the annotations for 9 generative factors. Since we know in advance the number and dimensions of the dataset's generative factors, as reported in Table~\ref{tab:dsents_factors}, we set our DCTC model to consider that specific setting, namely 9 discrete generative factors with their true dimensions. The parameter $\gamma$ from Fig.~\ref{eq:elbo_dec} is fixed at 50 while the discrete channel $C_{z_d}$ is increased from 0 to 30 in 25k steps. Similarly, we set the continuous models to learn 9 Gaussian latent variables, however, differently from the discrete case, it is not possible to map the continuous latent to a dimension.

% dsents factors
\begin{table}[t]
    \small \setlength\tabcolsep{4.5pt}
    \centering
    \begin{tabular}{|l|l|l|}
        \hline
        Factor         & Dimensions & Values                       \\ \hline
        Verb/object    & 1100       & [Verb/obj variations]        \\ \hline
        Gender         & 2          & [Male, Female]               \\ \hline
        Negation       & 2          & [Affirmative, Negative]       \\ \hline
        Tense          & 3          & [Present, Future, Past]        \\ \hline
        Subject number & 2          & [Singular, plural]           \\ \hline
        Object number  & 2          & [Singular, plural]           \\ \hline
        Sentence Type  & 2          & [Interrogative, Declarative] \\ \hline
        Person number  & 3          & [1st, 2nd, 3rd person]       \\ \hline
        Verb style     & 2          & [Gerund, Infinitive]         \\
        \hline
    \end{tabular}
    \caption{Generative factors in dSentences data.} \label{tab:dsents_factors}
\end{table}

\paragraph{Baselines.}
We compare our model against several types of state-of-the-art VAEs. 1) Continuous disentanglement models, such as~$\beta$VAE~\citep{higgins2016beta}, CCI-VAE~\citep{burgess2018understanding}, FactorVAE~\citep{kim2018disentangling} and $\beta$TC-VAE~\citep{chen2018isolating}. 2) Discrete VAEs, such as JointVAE~\citep{dupont2018learning} and VQVAE~\citep{van2017neural}. The JointVAE is a disentanglement method that jointly factorizes discrete and continuous variables, while VQVAE encodes discrete variables but does not encourage disentanglement. 3) Disentanglement models for text, such as Controlled Generation of Text (CGT)~\citep{hu2017toward}, which disentangles the sentence style, and Style Transfer VAE (ST-VAE)~\citep{john2019disentangled}, that disentangles style from content.

\newcommand{\ckm}{\textcolor{green!60!black}{\checkmark}}

\begin{table}[t]
    \small \centering
    \begin{tabular}{l cccccc|ccc}
        \hline
        & \multicolumn{6}{c}{Continuous} & \multicolumn{3}{c}{Discrete} \\ \hline
                   & 1    & 2    & 3    & 4    & 5    & 6    & 7    & 8    & 9    \\ \hline

        Verb tense & x    & \ckm & x    & \ckm & x    & x    & x    & x    & \ckm \\
        Subj-num   & x    & \ckm & \ckm & \ckm & \ckm & \ckm & \ckm & \ckm & \ckm \\
        Obj-num    & x    & \ckm & \ckm & \ckm & \ckm & \ckm & \ckm & \ckm & \ckm \\
        Gender     & \ckm & \ckm & \ckm & \ckm & x    & x    & x    & \ckm & \ckm \\
        Sent-type  & \ckm & x    & \ckm & \ckm & x    & x    & x    & x    & \ckm \\
        Person-num & x    & x    & \ckm & x    & x    & x    & x    & x    & \ckm \\
        Verb-style & x    & x    & x    & x    & x    & x    & x    & x    & x    \\
        Negation   & \ckm & \ckm & x    & \ckm & \ckm & \ckm & x    & \ckm & \ckm \\
        Verb/obj   & \ckm & \ckm & \ckm & \ckm & x    & x    & x    & x    & \ckm \\
        \hline
    \end{tabular}
    \caption{Summary of traversal for all latents. Models are abbreviated (1=$\beta$VAE, 2=CCIVAE, 3=$\beta$TCVAE, 4=FactorVAE, 5=CGT, 6=STVAE, 7=VQVAE, 8=JointVAE, 9=DCTC).
    }
    \label{tab:trav_results}
\end{table}

\paragraph{Traversal Analysis.}
% table settings
We consider a traversal for a factor to be disentangled only if the decoded sentences assume different values for the traversed factor, keeping all other factors unchanged, as explained in the previous section. The results for all generative factors are reported in Table~\ref{tab:trav_results}, where the checkmark and cross symbolise respectively disentangled and entangled factors.

% results
We see that DCTC achieves the most consistent semantics, by correctly disentangling 8 out of 9 generative factors. $\beta$TC-VAE and FactorVAE are able to disentangle respectively 6 and 7 out of 9 factors, while the NLP model (i.e.\ CGT and ST-VAE) are outperformed and only disentangle 3 out of 9 factors. We attribute the success of DCTC to the fact that it is modeling factors knowing their dimensions, and encourages disentanglement on the latents, after having encoded them as discrete variables. On the other hand, encoding discrete variables does not guarantee disentanglement by itself, as shown by the poor performance of VQVAE, while only accounting for style and content embeddings also leads to entangled representations, as shown by CGT and ST-VAE.

%We finally note that factors such as subject and object numbers result disentangled by most models, while verb-style remains entangled in all models. We hypothesize that disentangling verb-style is challenging due to its similarity with tense.

% examples
Samples of decoded sentences from the traversals for tense and subject-number are displayed in Table~\ref{tab:trav_examples}. DCTC correctly disentangles the tense into present, future and past, while the other factors are fixed, and similarly it disentangles the subject-number factor. On the other hand, $\beta$-VAE and JointVAE show an entangled representation for tense and subject-number. 

% The $\beta$-VAE has fixed values for tense, and it wrongly  disentangles verb/object, and also fails to disentangle subject-number. The JointVAE disentangles tense, but it also changes the verb/object factor, and change the subject-number to the wrong person (2nd instead of 1st).

% traversals results

\begin{table}[t]
\scriptsize

\begin{center}
\begin{tikzpicture}
\node[fill=lightgray] (table) [inner sep=2.5pt] {
\begin{tabular}{ ll | l}
& Tense & Subject-number \\ \hline
input      & you \textcolor{teal}{will} not attend the party     & \textcolor{teal}{we} will not attend the party \\  \hline
& & \\

$\beta$VAE & you \textcolor{blue}{will} not attend the party      & we will not attend the party    \\
& you \textcolor{red}{will} not \textcolor{red}{sign the paper}   & \textcolor{red}{he} will not attend the party \\
& you \textcolor{red}{will} not attend the party                  &  \\
& & \\

JointVAE   & you \textcolor{blue}{will} not attend the party      & we will not attend the party                   \\
& you \textcolor{blue}{did} not \textcolor{red}{join the wedding} &  \textcolor{red}{you} will not attend the party\\
& you \textcolor{blue}{do} not attend the party                   &  \\
& & \\

DCTC       & you \textcolor{blue}{will} not attend the party     & we will not attend the party                   \\
& you \textcolor{blue}{did} not attend the party                  & \textcolor{blue}{i} will not attend the party \\
& you \textcolor{blue}{do} not attend the party                  &  \\

\end{tabular}
};
\draw [rounded corners=.5em, line width=1.5pt] (table.north west) rectangle (table.south east);
\end{tikzpicture}
\end{center}

\caption{Traversal examples for tense and subject-number. Disentangled factors in \textcolor{blue}{blue}, entangled in \textcolor{red}{red}.}
\label{tab:trav_examples}
\end{table}

\subsection{Quantitative Evaluation}

\paragraph{Measuring Text Disentanglement.}
Quantifying disentanglement is a necessary step in our evaluation, in order to make the qualitative assessment more granular. Current disentanglement metrics in Representation Learning~\citep{higgins2016beta,kim2018disentangling}, rely on image-based datasets that provide the true generative factors, however, most datasets do not present such annotations. Fortunately, text data has the advantage, over images, of being discrete and regular by nature, and thus, generative factors can be defined at a sentence-level, by leveraging linguistically robust syntactic and semantic categories. Following this intuition, a simple solution for being able to measure disentanglement in a text representation, is to have a pre-processing step, where generative factors are extracted, before utilizing this information to compute the quantitative metrics.

\paragraph{Data Preparation.}
In our experiment, we focus on the Yelp reviews dataset~\citep{shen2017style}, which is composed by 600,000 review sentences, and we define and extract 5 generative factors, namely: gender, tense, negation, subject number, and object number. Using the part-of-speech (POS) engine provided by the Stanza python package~\citep{qi2020stanza}, we extract: 1) the gender factor from the pronouns and 2) the number factor from subjects and objects. Similarly, the tense is obtained from the verb using Stanza's lemmatizer, while  negation is determined from the presence of negation attributes in the parsed metadata.

\paragraph{Experimental Setup.}
We follow the setup of the previous qualitative experiment in terms of baselines and parameters. We investigate the models on two datasets, namely, Yelp, where 5 generative factors are extracted with the described data preparation process, and dSentences, where 9 factors are provided. As a result, we set the DCTC model with respectively 5 and 9 latent variables, and in both cases we use the dimensions for linguistic features defined in Table~\ref{tab:dsents_factors}.

\paragraph{Evaluation Metrics.}
We note that current disentanglement approaches in NLP~\citep{john2019disentangled} measure factors such as style transfer strength, content preservation, and quality of generation. In contrast, we are interested in evaluating the representations by computing the amount of disentanglement in the latents. To this end, we leverage the generative factors extracted from raw text, and compute three disentanglement metrics from the Representation Learning literature, namely \textit{Z-diff}~\citep{higgins2016beta}, \textit{Z-min-var}~\citep{kim2018disentangling}, and Mutual Information Gap (MIG)~\citep{chen2018isolating}. The main difference between Z-diff, Z-min-var and MIG is that the first two are reporting the accuracy of a classifier trained to recognize disentangled factors, while the last one is centered on measuring mutual information between latents and observed variables. More details about the metrics are reported in the appendix.

\paragraph{Metrics Analysis.}
The quantitative results in Table~\ref{tab:quant_evaluation} show that DCTC outperforms the other baselines in 2 out of 3 metrics, for both datasets. Specifically, DCTC achieves the best performance in terms of MIG and Z-min-var, while obtaining the second best scores for the Z-diff metric after JointVAE and FactorVAE, respectively on the Yelp and dSentences datasets. These measurements are overall confirming the hypothesis of the qualitative experiment, namely, that the proposed DCTC model is able to achieve a disentangled representation of language, by jointly optimizing independence of latents and accounting for the discrete nature of the data.

Another similarity with the qualitative experiment is that both VQVAE and the continuous NLP baselines (CGT and ST-VAE) are on average outperformed by other models. The outcome of VQVAE indicates that despite the fact that discrete variables can encode the dimensions of linguistic features, a model should also encourage independence of the variables, in order to achieve competitive scores in disentanglement metrics. Similarly, we hypothesize that both CGT and ST-VAE do not necessarily consider all the generative factors that the experiment is evaluating, in fact, CGT only aims to disentangle the style embedding of sentences, while ST-VAE focuses on disentangling style from content.

%disentanglement metrics results

\begin{table}[t]
    \small \centering
    \setlength\tabcolsep{3.5pt}
    \begin{tabular}{ clccc|ccc}
        \hline
        & & \multicolumn{3}{c|}{dSentences} & \multicolumn{3}{c}{Yelp} \\ \hline
        &               & Z-min          & Z-diff         & MIG            & Z-min          & Z-diff        & MIG            \\ \hline
        \multirow{4}{*}{\rotatebox[origin=c]{90}{Cont.}}
        & CCI-VAE       & 0.79           & 0.71           & 0.23           & 0.83           & 0.84          & 0.25           \\
        & $\beta$VAE    & 0.88           & 0.87           & 0.32           & 0.91           & 0.75          & 0.30           \\
        & $\beta$TC-VAE & 0.92           & 0.90           & 0.27           & 0.91           & 0.92          & 0.29           \\
        & FactorVAE     & 0.91           & \textbf{0.92 } & 0.18           & 0.92           & 0.92          & 0.27           \\
        & CGT           & 0.78           & 0.63           & 0.13           & 0.77           & 0.66          & 0.18           \\
        & ST-VAE        & 0.82           & 0.67           & 0.24           & 0.84           & 0.72          & 0.26           \\ \hline
        \multirow{3}{*}{\rotatebox[origin=c]{90}{Disc.}}
        & VQVAE         & 0.77           & 0.74           & 0.27           & 0.75           & 0.76          & 0.17           \\
        & JointVAE      & 0.89           & 0.81           & 0.35           & 0.90           & \textbf{0.95} & 0.33           \\
        & DCTC          & \textbf{0.94 } & 0.91           & \textbf{0.43 } & \textbf{0.94 } & 0.92          & \textbf{0.49 } \\
        \hline

    \end{tabular}
    \caption{Disentanglement metrics results.} \label{tab:quant_evaluation}
\end{table}

\subsection{Text Style Transfer}

\paragraph{Arithmetics for Latent Factors.}
In this experiment, we take inspiration from previous work from \citet{mikolov2013linguistic}, which showed that word embeddings can capture semantic relations via vector arithmetics, (for example, king - man + woman = queen). More specifically, we consider text generative factors (e.g.\ negation), and investigate sentence-level embeddings arithmetic in the task of text style transfer (extrinsic evaluation of the models).

The style transfer protocol of our experiment is performed as follows. We first select a factor (e.g.\ negation) and extract two lists of sentences containing two specific values (e.g.\ negative, and affirmative), that we name respectively $s_n$ and $s_a$. The extraction is performed based on the generative factors that we obtained with the pre-processing from our previous experiment. We then compute the vector of the arithmetic difference between the latents of the two vectors, namely $v=s_a - s_n$. Intuitively, this operation removes the negative components from the positive vector. Finally, we consider a third vector of negative sentences, encode them to obtain the embedding, and sum the previously computed vector $v$. After decoding we expect the sentences to be characterised by an affirmative style.

\paragraph{Experimental Setup.}
We follow our quantitative experiment for the model setup, and datasets. We compare our model with 3 state-of-the-art style transfer models, including: iVAE~\citep{fang2019implicit}, DAAE~\citep{shen2020educating}, ST-VAE, along with top performing models from the previous experiments, namely FactorVAE, $\beta$TC-VAE, JointVAE. In terms of evaluation metrics, we report the style transfer accuracy for each factor. The accuracy is computed by extracting the factors using the same procedure used for their selection.

% style transfer results

\begin{table}[t]
    \small
    \centering
    \begin{tabular}{ clccccc  }
        \hline
        &               & \scriptsize{Gender} & \scriptsize{Negation} & \scriptsize{Tense} & \scriptsize{Subj}
        & \scriptsize{Obj}
        \\ \hline
        & & \multicolumn{5}{c}{dSentences} \\ \hline
        \multirow{6}{*}{\rotatebox[origin=c]{90}{Cont.}}
        & FactorVAE     & 0.85                & 0.78                  & 0.71               & 0.81              & 0.76             \\
        & $\beta$TC-VAE & 0.70                & 0.76                  & 0.45               & 0.75              & 0.81             \\
        & ST-VAE        & 0.79                & 0.81                  & 0.56               & 0.79              & 0.92             \\
        & iVAE          & 0.82                & 0.85                  & 0.56               & \textbf{0.90}     & 0.88             \\
        & DAAE          & 0.89                & 0.93                  & 0.56               & 0.89              & 0.85             \\
        \hline
        \multirow{2}{*}{\rotatebox[origin=c]{90}{Disc.}}
        & JointVAE      & 0.72                & 0.83                  & 0.67               & 0.87              & 0.87             \\
        & DCTC          & \textbf{0.90}       & \textbf{0.94}         & \textbf{0.73}      & 0.86              & \textbf{0.95}    \\
        \hline
        & & \multicolumn{5}{c}{Yelp} \\ \hline
        \multirow{6}{*}{\rotatebox[origin=c]{90}{Cont.}}
        & FactorVAE     & 0.83                & 0.89                  & 0.23               & 0.72              & 0.80             \\
        & $\beta$TC-VAE & 0.67                & 0.72                  & 0.47               & 0.81              & 0.78             \\
        & ST-VAE        & 0.71                & 0.93                  & 0.43               & 0.83              & 0.89             \\
        & iVAE          & 0.85                & 0.82                  & 0.61               & 0.79              & 0.90             \\
        & DAAE          & 0.83                & 0.92                  & 0.52               & \textbf{0.89}     & \textbf{0.92}    \\
        \hline
        \multirow{2}{*}{\rotatebox[origin=c]{90}{Disc.}}
        & JointVAE      & 0.81                & 0.80                  & 0.59               & 0.80              & 0.80             \\
        & DCTC          & \textbf{0.89}       & \textbf{0.96}         & \textbf{0.65}      & 0.87              & 0.87             \\   \hline

    \end{tabular}
    \caption{Style Transfer Accuracy.}
    \label{tab:transfer_accuracy}
\end{table}

%        & $\beta$VAE    & 0.83                & 0.89                  & 0.50               & 0.78              & 0.83             \\
%    & VQVAE  & 0.88      & 0.89        & 0.39     & 0.79      & 0.75      & 0.83       & 0.45          & 0.72          \\
%        & $\beta$VAE    & 0.85                & 0.78                  & 0.34               & 0.76              & 0.86             \\

\paragraph{Style Transfer Analysis.}
The style transfer accuracy results reported in Table~\ref{tab:transfer_accuracy} shows that our DCTC model outperforms other baselines for the majority of the factors, for both Yelp and dSentences. DCTC achieves achieves the second best result for subject-number after iVAE and DAAE, respectively on dSentences and Yelp.

We hypothesize that DCTC achieves the strongest performance due to the fact that it is explicitly set to learn variables with a known dimension, which can not be achieved with continuous models. Furthermore, by disentangling the latent variables, DCTC is able to provide a representation that results suitable for the task of flipping single factors. Finally, the strength of disentanglement is also highlighted by the fact that $\beta$TC-VAE and FactorVAE are performing comparably with the style transfer models, even if they are not created originally for this task.

Some examples for style transferred sentences from Yelp are reported in Table~\ref{tab:transfer_examples}. We can observe that some baselines are able to invert the considered factors (tense and negation), however, DCTC is the only one that correctly inverts both factors without the need of changing other words from the input sentence. This can be justified by the ability of disentangled representation to encode invariance of certain factors. This extrinsic evaluation confirms the hypothesis that the joining disentangled representation with discrete encoding can positively impact the downstream task and represents a fundamental tool to design more expressive language encodings.

\begin{table}
\scriptsize

\begin{subtable}{0.48\textwidth}
\begin{center}
\begin{tikzpicture}
\node[fill=lightgray] (table) [inner sep=5.5pt] {
\begin{tabular}{l p{5.5cm}}
input         & the pizza served \textcolor{teal}{was} missing a portion    \\ \hline
DAAE          & the pizza \textcolor{blue}{is} served a little bland         \\
iVAE          & the pizza served \textcolor{blue}{is} for a ridiculous price \\
$\beta$VAE    & the pizza \textcolor{red}{was} mistaken                      \\
$\beta$TC-VAE & the pizza \textcolor{blue}{is} completely wrong!             \\
ST-VAE        & the pizza served \textcolor{blue}{is} pretty decent          \\
JointVAE      & our pizza \textcolor{red}{was} great                         \\
DCTC          & the pizza served \textcolor{blue}{is} missing a portion      \\
\end{tabular}
};
\draw [rounded corners=.5em, line width=1.5pt] (table.north west) rectangle (table.south east);
\end{tikzpicture}
\begin{tikzpicture}
\node[fill=lightgray] (table) [inner sep=5.5pt] {
\begin{tabular}{l p{5.5cm}}
input         & he told me he could \textcolor{teal}{not} exchange them   \\ \hline

DAAE          & he told me he \textcolor{blue}{could} exchange them        \\
iVAE          & he told me he \textcolor{blue}{could} be more attentive    \\
$\beta$VAE    & he told me that he does \textcolor{red}{not} exchange them \\
$\beta$TC-VAE & he told me he \textcolor{blue}{wanted} a second opinion    \\
ST-VAE        & he told me to \textcolor{blue}{try} the plantains          \\
JointVAE      & he told us he could \textcolor{red}{not} be happier        \\
DCTC          & he told me he \textcolor{blue}{could} exchange them        \\
\end{tabular}
};
\draw [rounded corners=.5em, line width=1.5pt] (table.north west) rectangle (table.south east);
\end{tikzpicture}

\end{center}

\end{subtable}
\caption{Style transfer on Yelp. Top: tense, past to present. Bottom: negation, negative to affirmative. Correct changes in \textcolor{blue}{blue}, wrong ones in \textcolor{red}{red}.}
\label{tab:transfer_examples}
\end{table}

\section{Conclusion}
In this work, we propose the first approach where a discrete encoding of the linguistic features in sentences is integrated  with an objective function that encourages disentanglement. We provide a VAE-based architecture where latent variables are back-propagated with a discrete reparameterization mechanism. We then design a decomposition of the ELBO, where 1) the independence between latent variables is encouraged, to aid disentanglement, and 2) the amount of encoded information is controlled, to avoid the posterior collapse. We provide a novel evaluation procedure where representations learned from text data are probed in terms of their disentanglement, using metrics from Representation Learning. With this evaluation tool, we demonstrate that the presented model consistently outperforms continuous and discrete baselines for disentanglement, on qualitative evaluation, quantitative metrics, and text style transfer.

We conclude that the modeling of discrete variables, which is currently under-explored in disentanglement research, may represent a fundamental encoding tool for enhancing interpretability and control in NLP models.

\section*{Acknowledgements}
The research was partially funded by EPSRC and the BBC under iCASE and EPSRC Enncore (EP/T026995/1). The authors would like to thank Marco Valentino and Mokanarangan Thayaparan for the helpful discussions and Chris Newell for his project support.

\bibliographystyle{acl_natbib}
\bibliography{references}

\newpage

\appendix

\section{Disentanglement Metrics}
\paragraph{Z-diff}
Z-diff~\citep{higgins2016beta} considers pairs of instances to create batches where a generative factor $k$ is chosen randomly. We then consider the pairs $v_1$ $v_2$ that have the same value for the factor $k$. The absolute difference of the encoding of the pair is then determined, namely $|z_1 - z_2|$. The intuition of the metric is that a smaller difference for a fixed factor entails more similar samples. A reference dataset is constructed with this procedure, then a linear classifier is trained to predict which factor is fixed, and the accuracy is considered as the disentanglement metric.

\paragraph{Z-min-var}
The Z-min-var~\citep{kim2018disentangling} is similar to the Z-diff metric, as it creates a reference dataset and train a classifier to find the fixed factor. Specifically, Z-min-var builds the dataset using the $argmin$ of the variance vector of all encodings, which have been normalized by the standard deviation.

Z-diff and Z-min-var both rely on the intuition that a smaller difference for a fixed factor entails more similar samples. They create a dataset for a classifier to predict the fixed factor, the accuracy of which is a measure for disentanglement. Z-diff creates the dataset with the absolute differences of encoding pairs where a factor is fixed, while Z-min-var builds a similar dataset, but using the $argmin$ of the variance vector of all encodings, which have been normalized by the standard deviation.

\paragraph{Mutual Information Gap}
 MIG~\citep{chen2018isolating} does not rely on a classifier, and thus it provides more robustness against hyperparameter biases. MIG first computes the mutual information between each latent and the true factor, and then it identifies and subtracts the two values for latents with maximum mutual information. The obtained quantity is considered the amount of disentanglement.

\end{document}